\documentclass[conf]{new-aiaa}
\usepackage[utf8]{inputenc}

\usepackage{graphicx,subfig}
\usepackage{amsmath}
\usepackage[version=4]{mhchem}
\usepackage{siunitx}
\usepackage{longtable,tabularx}
\usepackage[table,xcdraw]{xcolor}
\usepackage{graphicx}
\usepackage[normalem]{ulem}
\useunder{\uline}{\ul}{}
\usepackage{wrapfig}
\usepackage{float} 

\title{Integrating Vision Systems and STPA for Robust Landing and Take-Off in VTOL Aircraft}

\author{
    Sandeep Banik\footnote{Postdoctoral Research Associate, Department of Mechanical Science and Engineering, University of Illinois Urbana-Champaign, Urbana, IL 61801, USA},
    Jinrae Kim\footnote{Postdoctoral Research Associate, Department of Mechanical Science and Engineering, University of Illinois Urbana-Champaign, Urbana, IL 61801, USA}
    and Naira Hovakimyan\footnote{W. Grafton and Lillian B. Wilkins Professor, Department of Mechanical Science and Engineering, University of Illinois Urbana-Champaign, Urbana, IL 61801, USA. AIAA fellow}
}
\affil{Department of Mechanical Science and Engineering, University of Illinois Urbana-Champaign, Urbana, IL 61801, USA}
\author{
    Luca Carlone\footnote{Boeing Career Development Associate Professor, Department of Aeronautics and Astronautics, Massachusetts Institute of Technology, Cambridge, MA 02139, USA},
    John P. Thomas\footnote{Research Engineer, Department of Aeronautics and Astronautics, Massachusetts Institute of Technology, Cambridge, MA 02139, USA}
    and Nancy G. Leveson\footnote{Jerome C. Hunsaker Professor, Department of Aeronautics and Astronautics, Massachusetts Institute of Technology, Cambridge, MA 02139, USA}
}
\affil{Department of Aeronautics and Astronautics, Massachusetts Institute of Technology, Cambridge, MA 02139, USA}

\begin{document}

\maketitle

\section*{Abstract}
\textbf{
Vertical take-off and landing (VTOL) unmanned aerial vehicles (UAVs) are versatile platforms widely used in applications such as surveillance, search and rescue, and urban air mobility. Despite their potential, the critical phases of take-off and landing in uncertain and dynamic environments pose significant safety challenges due to environmental uncertainties, sensor noise, and system-level interactions. This paper presents an integrated approach combining vision-based sensor fusion with System-Theoretic Process Analysis (STPA) to enhance the safety and robustness of VTOL UAV operations during take-off and landing. By incorporating fiducial markers, such as AprilTags, into the control architecture, and performing comprehensive hazard analysis, we identify unsafe control actions and propose mitigation strategies. Key contributions include developing the control structure with vision system capable of identifying a fiducial marker, multirotor controller and corresponding unsafe control actions and mitigation strategies. The proposed solution is expected to improve the reliability and safety of VTOL UAV operations, paving the way for resilient autonomous systems.}

\section{Introduction}
Vertical take-off and landing (VTOL) unmanned aerial vehicles (UAVs) have emerged as a vital technology in various applications~\cite{wei2024autonomous}, including surveillance, search and rescue, and urban air mobility~\cite{silva2018vtol}. Their ability to hover and operate in confined spaces without requiring runways~\cite{simmons2021full} makes them highly versatile. However, the critical phases of take-off and landing in uncertain and dynamic environments pose significant challenges. Environmental uncertainties, sensor noise, and model inaccuracies can adversely affect the UAV's performance, leading to potential safety hazards and mission failures~\cite{xia2022dynamics,he2022adaptive}. Ensuring robust autonomous operations under these uncertainties is essential for the broader adoption of VTOL UAVs. Vision-based systems have been identified as a promising solution to enhance situational awareness, enable precise navigation and take-off and land efficiently by providing rich environmental information~\cite{achtelik2009stereo}. Specifically, integrating vision systems like AprilTag detection can significantly improve landing pad identification and tracking, which is crucial during take-off and landing phases~\cite{olson2011apriltag}.

While incorporating various components can enhance the capabilities of a system, it also introduces significant complexity in analyzing and assuring its safety. Addressing safety concerns for a whole system requires a systematic approach to hazard analysis and develop corresponding mitigation strategies. When independently designed components are integrated, their interactions under complex dynamics and environmental factors—such as wind gusts or varying lighting conditions during landing zone detection can degrade system performance or even lead to safety violations~\cite{gautam2014survey,alam2021survey}.
For instance, numerous learning-based vision algorithms exist for tasks such as navigation, scene understanding, and obstacle avoidance. However, these algorithms are often tailored for specific purposes and lack a holistic approach to system safety, particularly in addressing hazards that emerge from interactions between different components. The System-Theoretic Process Analysis (STPA) framework offers a comprehensive method to identify potential hazards, unsafe control actions, and design constraints in complex systems~\cite{leveson2016engineering}. By incorporating STPA into the VTOL UAV's control system development, we are able to proactively address safety issues arising from both system components and interactions. 

\begin{wrapfigure}{r}{0.45\textwidth} 
    \centering
    \includegraphics[width=0.42\textwidth]{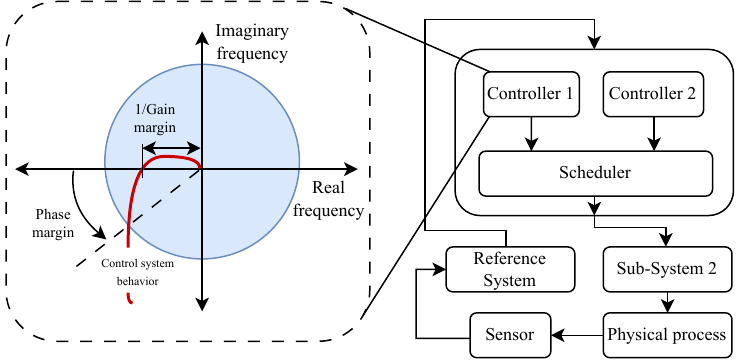} 
    \caption{\small Illustration of an arbitrary control structure~\cite{leveson2016engineering}, with an expanded view of controller 1 performance. The dotted box expanded from Controller 1 represents a Nyquist plot with a gain and phase margin, indicating the performance of the controller over the frequency domain.}
    \label{fig:performance_to_safety}
\end{wrapfigure}
The motivation for this work arises from the need to connect performance of low-level components with the system-level safety requirement. 
An illustration of an arbitrary control structure (Step 2 of STPA~\cite{leveson2016engineering}) is shown in~\autoref{fig:performance_to_safety}. The illustration demonstrates the performance of controller 1; however, such a performance is not guaranteed when it is integrated with other sub-systems, which makes up the whole system. Components of complex systems are generally developed in parallel, often in a process that fails to fully capture the interdependencies necessary to meet high-level requirements. Although a controller may assure performance under specific conditions, these assurances often rely on underlying assumptions and constraints. A top-down system-theoretic approach enables us to analyze such interconnections to generate system and controller constraints, which connects the individual performance of sub-systems to the safety requirements. 

In this paper, we present an integrated approach that combines a vision-based sensor fusion mechanism with an autopilot of a VTOL UAV, and employ STPA for hazard analysis to improve the robustness during take-off and landing. The key contributions of this work include:
\begin{itemize}
    \item Development of a vision-based control structure and analysis: We identify a control structure that integrates an autopilot with a fiducial marker system, such as AprilTag detection.
    We expand the component blocks in a cascading manner until we reach the control structure with the vision processing block.
    A comprehensive STPA is conducted on this control structure to identify unsafe control actions and scenarios under which the safety of the VTOL UAV could be compromised. 
    \item Mitigation strategies for vision-based and multirotor control structures: We propose a set of effective mitigation strategies tailored for the vision-based and multirotor control structures, ensuring the safety of VTOL UAVs during take-off and landing operations. 
\end{itemize}
This integrated approach addresses the safety concerns of a VTOL UAV with a vision-based system, while operating in autonomous mode. The results demonstrate the gap in assuring safety requirements when integrating different components in a VTOL, and recommend mitigation strategies to improve the robustness and reliability during critical flight phases, contributing to the advancement of VTOL UAV technologies.

\section{Literature Survey}
VTOL UAVs have gained significant attention due to their wide-ranging applications, including urban air mobility, logistics, surveillance, and disaster response~\cite{smith2012projected,patterson2012performance,moore2014misconceptions,cohen2021urban}. VTOL UAVs are typically classified into three major configurations: tailsitter, tiltrotor, and tiltwing designs~\cite{ducard2021review,hochstenbach2015design} . Each configuration offers distinct advantages and trade-offs. Tailsitter UAVs transition from vertical take-off to horizontal flight by reorienting their entire body, while tiltrotor UAVs achieve transitions by tilting the rotors between vertical and horizontal positions. Similarly, tiltwing UAVs tilt the entire wing to ensure smoother transitions. These designs differ in terms of performance, stability, efficiency, and maneuverability, and their suitability depends heavily on the mission requirements, such as long-range endurance or short-distance agility~\cite{bacchini2019electric,saeed2015review}. With approximately 250 companies currently working to develop commercial VTOL UAVs, such as Joby Aviation, Lilium, Volocopter, Ehang, Airbus, Kittyhawk, and Wisk are leading the race to bring these advanced systems to the commercial space~\cite{ieee_evtoal_air_taxi}. 
Among the various VTOL designs, the tiltrotor configuration stands out as the most widely recognized for its aerodynamic efficiency, particularly during the transition from hovering to low-speed forward flight~\cite{ding2022aerodynamic}. This efficiency makes tiltrotor UAVs highly suitable for applications that require both vertical agility and horizontal range. In this work, we focus on control structure identification for tiltrotor VTOLs, addressing the unique challenges associated with their operational stability and transition dynamics.

Despite the rapid advancements in VTOL UAVs, several significant challenges~\cite{xiang2024autonomous} persist in achieving robust autonomy and reliable performance, namely, i) safety and reliability, ii) sensing and perception~\cite{kanellakis2017survey,araar2014new,korn2008uas}, iii) decision making~\cite{zulu2016review,boubeta2018autonomous}, iv) automated flight controller synthesis~\cite{afari2023review}, v) regulatory challenges~\cite{edwards2020evtol}, and iv) societal challenges. One of the most fundamental challenges is safety and reliability. VTOL UAVs are expected to meet stringent safety standards that surpass those of existing aviation platforms due to their complex multimodal operations and broader deployment potential. Safety concerns are deeply intertwined with regulatory and societal challenges, making safety a critical focus of development efforts. 

Sensing and perception represent another critical area for VTOL development, as autonomous operations rely heavily on accurate environmental awareness. These systems form the foundation for decision-making processes and the synthesis of automated flight controllers~\cite{kanellakis2017survey,araar2014new,korn2008uas}. Autonomous decision-making itself presents additional challenges, requiring algorithms capable of managing transitions between flight modes, navigating complex airspaces, and responding to dynamic environments~\cite{zulu2016review,boubeta2018autonomous}. Automated flight controllers further complicate development, as they must handle highly nonlinear dynamics while maintaining stability and efficiency under varying operational conditions~\cite{afari2023review}. Hardware-in-the-loop (HIL) simulations are often used to verify VTOL functionality and validate control performance~\cite{nielsen2021visually,kulkarni2022simulation}. However, these simulations frequently lack a comprehensive, top-down approach to safety assurance, highlighting the need for more advanced validation frameworks.


A promising approach to addressing safety challenges in VTOL UAVs is the application of System-Theoretic Process Analysis (STPA). STPA is a hazard analysis technique rooted in systems theory, designed to identify potential hazards by examining unsafe control actions within complex systems~\cite{leveson2016engineering}. Unlike traditional hazard analysis methods, STPA considers both component failures and unsafe interactions within the system, making it particularly well-suited for the intricate multimodal operations of VTOL UAVs.

STPA has been successfully applied in the aviation domain to analyze and mitigate risks in autonomous systems, including UAVs. For instance, STPA was applied to the Airworthiness Technologies Research Center UAV (ATRC-UAV)~\cite{chen2015stpa} to analyze take-off operations, focusing on a top-level control structure. In another study, multiple hazard analysis tools, such as Functional Hazard Analysis (FHA) and STPA, were combined to conduct a worst-case analysis of a VTOL carrying an urban air passenger~\cite{graydon2020guidance}. The safety analysis of VTOLs using STPA was further extended to include manufacturers and operators to support regulatory agencies in the certification process~\cite{Filho_STPA_2024}. Additionally, STPA has been applied to an existing design architecture of a VTOL to identify potentially unsafe issues and modify the architecture to enhance safety~\cite{ning2023safe}.

While prior works have primarily implemented STPA at a high-level control structure to identify generic unsafe control actions and provide safety recommendations, this work takes a different approach. Here, we expand the control structure to multiple levels, with a specific focus on vision-based systems, to identify precise control actions that could become unsafe and propose mitigation strategies to address them.

\section{System Description}~\label{sec:system_description}
The system under consideration is a generic VTOL UAV equipped with a vision system, when operating in an autonomous mission. An autonomous mission for a VTOL UAV typically consists of five stages: i) take-off in hover mode, ii) transition to fixed-wing mode, iii) waypoint navigation in fixed-wing mode, iv) back-transition to hover mode, and v) landing in hover mode. For the purposes of this paper, we limit our analysis to the take-off and landing phases in hover mode, as these phases are critical and prone to operational uncertainties. The VTOL UAV under consideration comprises the following key components:

\subsection{Host VTOL UAV}
The host VTOL UAV is assumed to have been developed independently, with the goal of creating a low-cost, collaborative platform capable of supporting a wide variety of autonomy systems. It operates under two primary control modes throughout a mission:
\begin{itemize}
    \item Control mode I: Remotely piloted by a human pilot on the ground (human-in-the-loop). 
    \item Control mode II: Piloted by an autopilot, where the remote human pilot may issue overriding commands to the autopilot (human-on-the-loop).
\end{itemize}
In Control mode I, the human pilot retains complete control of the UAV, managing all five stages of the mission. This mode requires constant visual feedback to ensure the effective execution of the mission. Conversely, Control mode II leverages the capabilities of the autopilot to perform all or parts of the mission autonomously. In this mode, the human pilot retains the ability to intervene and issue overriding commands to the autopilot, enabling corrective actions when needed. This dual-mode architecture ensures flexibility in operations, accommodating both manual control and varying levels of autonomy based on mission requirements.

\subsection{Autopilot}
Autopilot systems form the backbone of modern unmanned aerial vehicle (UAV) operations, enabling autonomous navigation and control through advanced algorithms and sensor integration. These systems perform critical functions, such as maintaining stability, executing predefined flight paths, and adapting to environmental changes with minimal human intervention. Similar to the host VTOL UAV, the autopilot is assumed to have been developed independently to ensure compatibility with a wide range of UAV systems. The autopilot comprises multiple components, including guidance logic, flight controllers (for both multi-rotor and fixed-wing configurations), tilt schedulers, and estimators. Widely used autopilot platforms, such as Ardupilot and PX4, provide open-source solutions that support diverse UAV configurations, including fixed-wing, rotary-wing, and hybrid designs. These platforms integrate data from various sensors, such as GPS for positional accuracy, inertial measurement units (IMUs) for attitude and orientation, and barometers for altitude control.

When an autopilot is integrated into a host VTOL UAV, the guidance logic and flight controllers must be tuned to enable seamless operation across different flight modes, including hover, transition, and fixed-wing configurations. This tuning process involves optimizing rotor tilt scheduling, critical tilt angles, and rotor wind-down rates controller gains~\cite{ducard2021review} to ensure accurate tracking of critical parameters such as attitude, altitude, and airspeed. 
By fine-tuning these parameters, the autopilot system significantly enhances the reliability and robustness of the VTOL UAV, enabling safe and efficient operations in diverse and unpredictable environments. While the autopilot includes multiple components, such as the guidance controller, flight controllers, and estimators, this paper refers to the autopilot specifically as the subsystem comprising the flight controllers.

\subsection{Ground Control Station}
The ground control station (GCS) serves as the primary interface between the remote human pilot and the VTOL UAV, enabling efficient mission planning, monitoring, and control. It provides the human pilot with real-time telemetry, including positional data, attitude information, battery status, and other critical system parameters. The GCS facilitates the configuration of flight plans, which include defining waypoints, altitude settings, and selecting flight modes.

In Control mode I, the GCS plays a pivotal role in providing the human pilot with visual feedback and situational awareness, which are essential for manual operation. Modern GCS platforms, such as QGroundControl and Mission Planner, are designed to be intuitive and user-friendly, supporting a wide range of autopilot systems to ensure seamless integration with various UAV configurations. Additionally, the GCS is equipped with data-logging capabilities, allowing for post-mission analysis and diagnostics. These capabilities are crucial for improving UAV performance, identifying system inefficiencies, and enhancing overall reliability.

\subsection{VTOL UAV Actuators}
The actuators and flight computer are essential components of the host VTOL UAV, enabling precise management of the vehicle's orientation, stability, and trajectory. The actuators, which may include rudders, elevators, ailerons, and tilt-rotors, are actuated by servos or motors. These actuators respond to commands generated by the flight controller based on real-time sensor feedback and guidance inputs, ensuring accurate control of the UAV's movements.



\begin{figure*}[ht]
	\begin{center}
		\subfloat[]{\includegraphics[width = 0.65\linewidth]{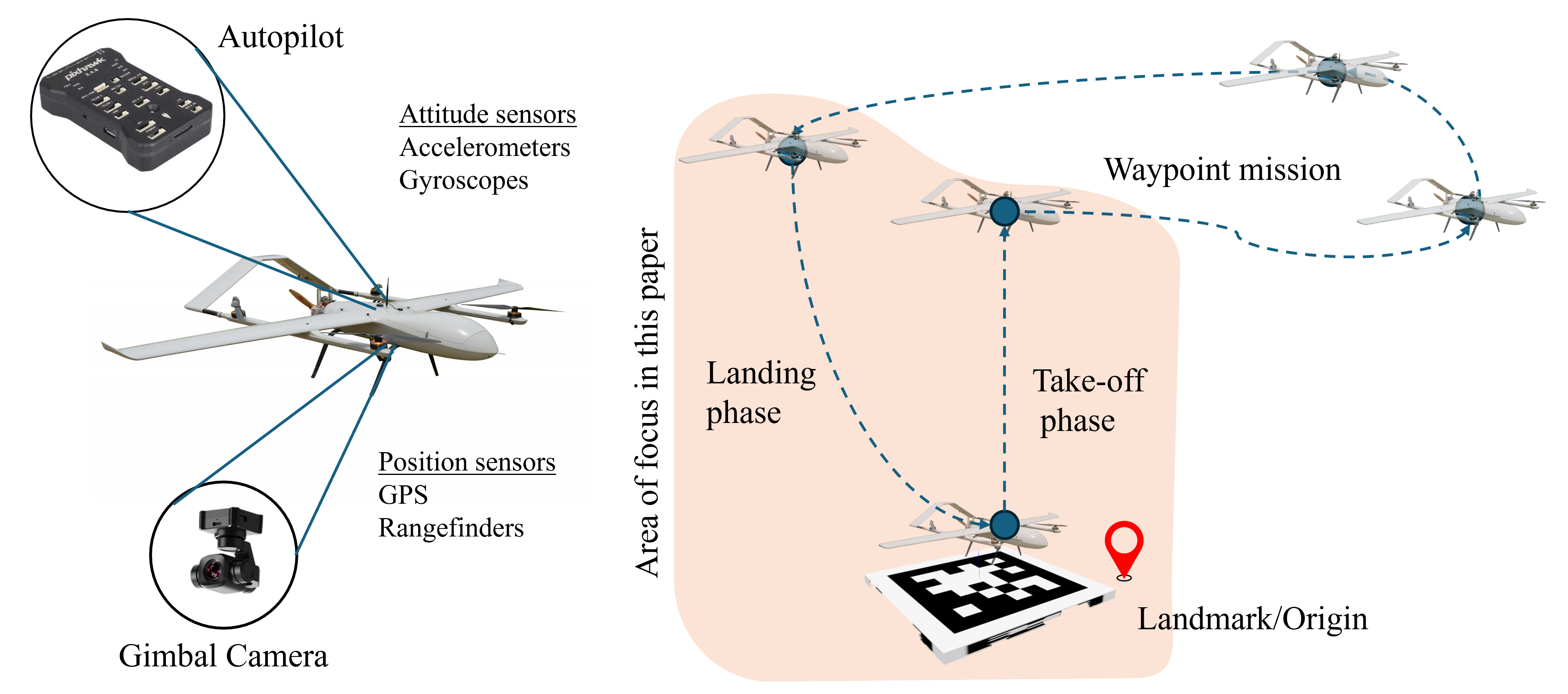}
			\label{fig:problem_illustration}	
		}
		\subfloat[]{\includegraphics[width = 0.30\linewidth]{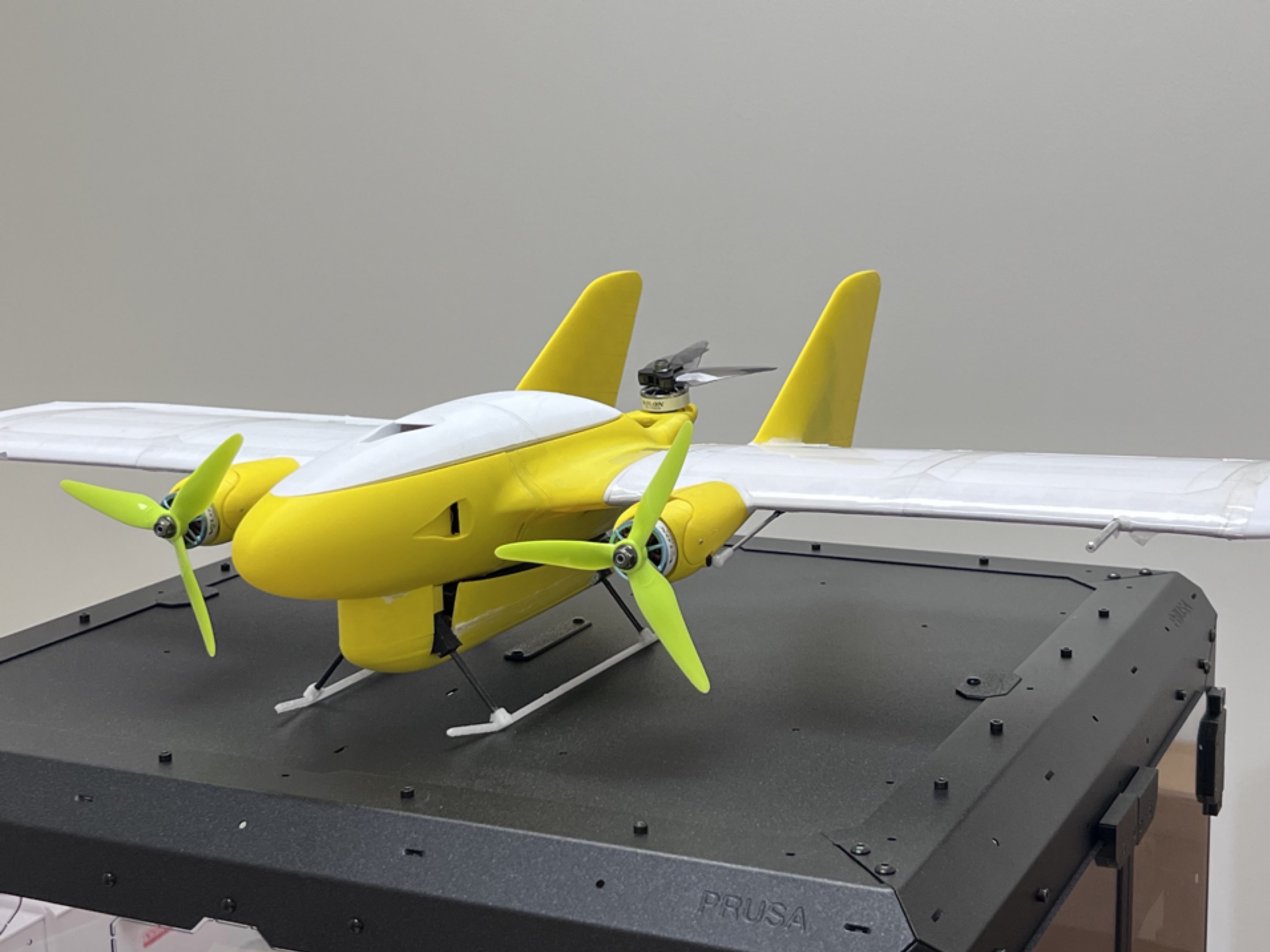}
			\label{fig:VTOL}	
		}
    \caption{\small (a) Illustration of the VTOL UAV mission phases, highlighting the take-off and landing phases as the focus of this study, with key onboard sensors and systems labeled. (b) Prototype of the VTOL UAV under development under University Leadership Initiative.}	
		\label{fig:Problem_description}
	\end{center}
\end{figure*}

\subsection*{Integration of VTOL UAV components}
As described earlier, the host VTOL UAV can be controlled either manually by a remote human pilot or autonomously via the autopilot. The human pilot issues control commands through an RC controller, which maps user inputs to specific control actions such as pitch, roll, yaw, and altitude adjustments. To initiate an autonomous mission, the pilot must explicitly enable the autonomy mode, which is disabled by default. Activating this mode triggers autonomous operations and communicates the status change to both the Ground Control Station (GCS) and the guidance controller.

In an autonomous mission, the human pilot defines mission parameters by specifying the launch location, waypoints, and the landing location (typically the same as the launch location). The GCS transmits this mission plan to the guidance controller. Once the autonomy mode is enabled, the UAV begins executing the mission. The guidance controller generates reference signals that the autopilot uses to ensure the UAV adheres to the predefined trajectory. The autopilot converts these reference signals into motor actuation and control surface commands, enabling precise maneuvering. A suite of onboard sensors—including accelerometers, magnetometers, barometers, GPS, and cameras continuously collects real-time flight data. Some of this sensor data are processed by the flight controller and relayed back to the different controllers in the autopilot block and guidance controller as feedback, closing the control loop. This integrated architecture facilitates robust control, real-time adaptation to environmental variations, and accurate execution of autonomous missions. By integrating STPA with vision-based capabilities, this study systematically identifies potential hazards, unsafe control actions, and critical safety constraints, ensuring robust operation even in complex and uncertain environments. This approach not only enhances the reliability of the UAV during take-off and landing but also establishes a foundation for designing resilient autonomy frameworks for future UAV systems. An illustration of the problem description is shown in Fig~\autoref{fig:problem_illustration} along with an ongoing VTOL development~\cite{carlson2021minihawk} at UIUC shown in Fig~\autoref{fig:VTOL}.


\section{System-Theoretic Process Analysis}

Safety of critical flight phases, such as take-off and landing, necessitates a structured and comprehensive approach to hazard identification and mitigation. System-Theoretic Process Analysis (STPA) provides a robust framework for analyzing complex systems by focusing on unsafe control actions and their impact on system safety~\cite{leveson2016engineering}. Unlike traditional hazard analysis methods that emphasize component-level failures, STPA adopts a system-wide perspective, considering interactions between components, environmental uncertainties, and dynamic operational contexts. This makes it particularly well-suited for addressing the challenges inherent in the vision-based control of VTOL UAVs.

Vision-based systems rely on high-resolution visual data to perform tasks such as landing pad detection, localization, and obstacle avoidance. However, the reliance on vision introduces unique challenges, including the possibility of degraded performance under adverse conditions, such as low light, occlusions, or sensor noise. STPA provides a structured methodology to analyze these potential failure modes and ensure safe operation by identifying hazards associated with vision system dependencies and their interactions with other UAV components.

The application of STPA to the take-off and landing phases involves the following steps, which guide the systematic identification and mitigation of hazards:

\subsection{Identification of losses and hazards}
The following losses were identified during the STPA process:
\begin{itemize}
    \item [L-1]: Loss or damage to host VTOL UAV or its environment. 
    \item [L-2]: Loss of credibility or reputation. 
    \item [L-3]: Loss of future test capability. 
    \item [L-4]: Inability to collect data for analysis.
\end{itemize}

Loss L-1 directly affects the host VTOL UAV and its operational environment, including critical assets such as the landing pad and nearby UAVs. Losses such as L-2 and L-3 have broader implications, particularly impacting the credibility and feasibility of deploying autonomous operations in the event of an accident. Lastly, loss L-4 pertains to scenarios where triggers prevent the collection of valuable data during autonomous operation tests. Addressing L-4 is crucial, as it supports the identification of risky evaluations and the development of improved safety mechanisms. For each identified loss, corresponding hazards and system-level constraints were determined, detailed in \autoref{tab:table_hazards} and \autoref{tab:STPA_system_constraints}, respectively.

\begin{table}[h]%
\caption{Identified hazards corresponding to the losses L-1 to L-4}
\centering
\begin{tabular}{clc}
\#  & \multicolumn{1}{c}{Hazards}                        & Connection to losses \\ \hline
H-1 & Host UAV too close to objects                      & {[}L-1, L-2, L-4{]}  \\
H-2 & Host UAV too close to other aircraft               & {[}L-1{]}-{[}L-4{]}  \\
H-3 & Host UAV departs authorized test area              & {[}L-1{]}-{[}L-4{]}  \\
H-4 & Host UAV lands outside the authorized landing area & {[}L-1{]}-{[}L-4{]}  \\
H-5 & Host UAV unable to provide useful test data        & {[}L-3,L-4{]}        \\
H-6 & Host UAV becomes uncontrollable                    & {[}L-1{]}-{[}L-4{]} 
\end{tabular}%
\label{tab:table_hazards}
\end{table}
\begin{table}[h]
\caption{Identified system-level constraints corresponding to hazards H-1 to H-6}
\resizebox{\columnwidth}{!}{%
\begin{tabular}{cll}
\# &
  \multicolumn{1}{c}{Hazards} &
  \multicolumn{1}{c}{System-level constraints} \\ \hline
H-1 &
  Host UAV too close to objects &
  \begin{tabular}[c]{@{}l@{}}SC-1: Host UAV must maintain TBD distance \\ from obstacles\end{tabular}
  \\
H-2 &
  Host UAV too close to other aircraft &
  \begin{tabular}[c]{@{}l@{}}SC-2: Host UAV must satisfy minimum \\ separation standard from other aircrafts\end{tabular} \\
H-3 &
  Host UAV departs authorized test area &
  \begin{tabular}[c]{@{}l@{}}SC-3: Host UAV must maintain operation\\ within the defined test area\end{tabular} \\
H-4 &
  Host UAV lands outside the authorized landing area &
  \begin{tabular}[c]{@{}l@{}}SC-4: Host UAV must land within \\ specified authorized landing area\end{tabular} \\
H-5 &
  Host UAV unable to provide useful test data &
  \begin{tabular}[c]{@{}l@{}}SC-5: Host UAV must maintain TBD seconds \\ of a mission to extract useful data\end{tabular} \\
H-6 &
  Host UAV becomes uncontrollable &
  \begin{tabular}[c]{@{}l@{}}SC-6: Host UAV must maintain controllability in\\  both or either autonomous or manual mode\end{tabular}
\end{tabular}%
}
\label{tab:STPA_system_constraints}
\end{table}
\begin{figure}[h!]
    \centering
    \includegraphics[width=0.60\linewidth]{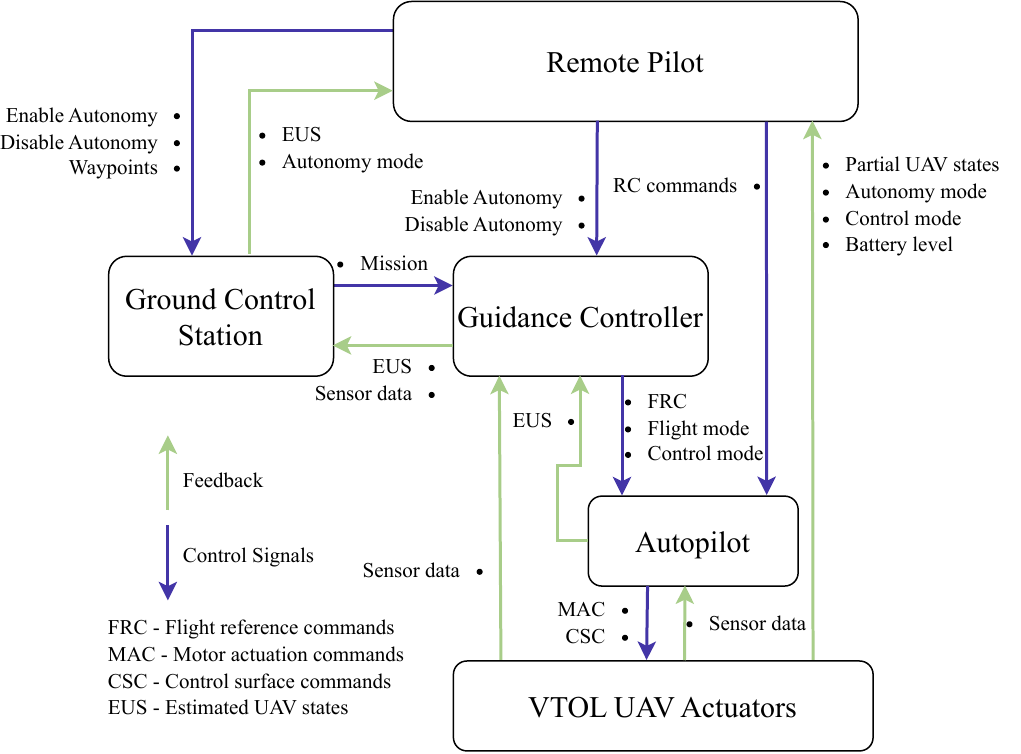}
    \caption{\small Control structure for the VTOL UAV.}
    \label{fig:Control_struct}
\end{figure}
\begin{figure*}[h!]
	\begin{center}
		\subfloat[]{\includegraphics[width = 0.49\linewidth]{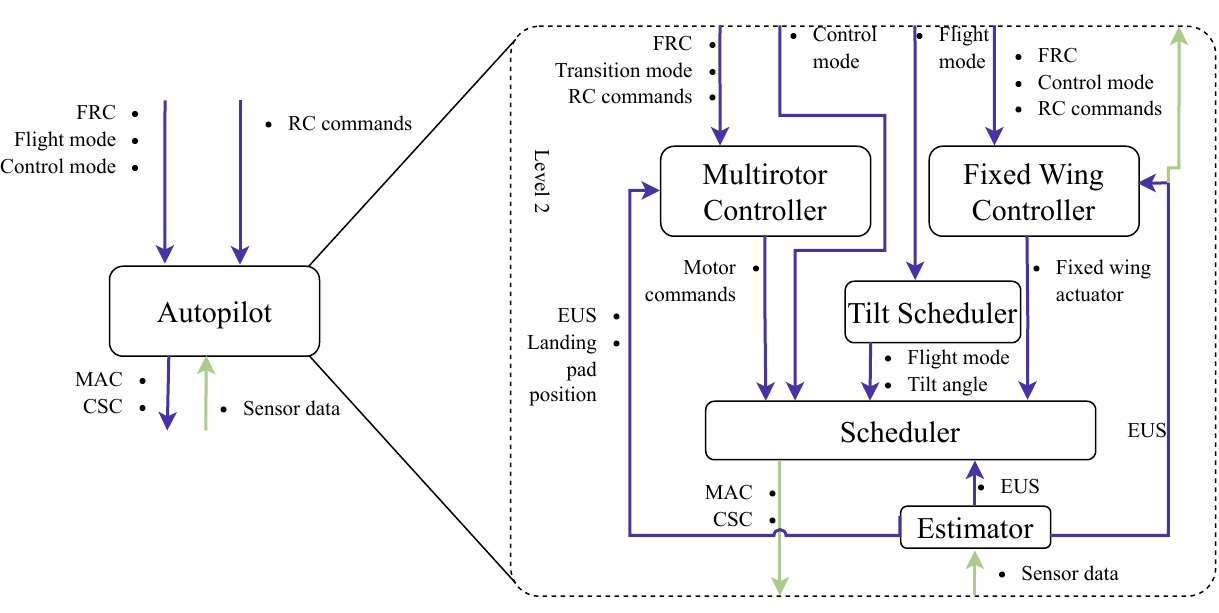}
			\label{fig:Control_struct_L2}	
		}
		\subfloat[]{\includegraphics[width = 0.49\linewidth]{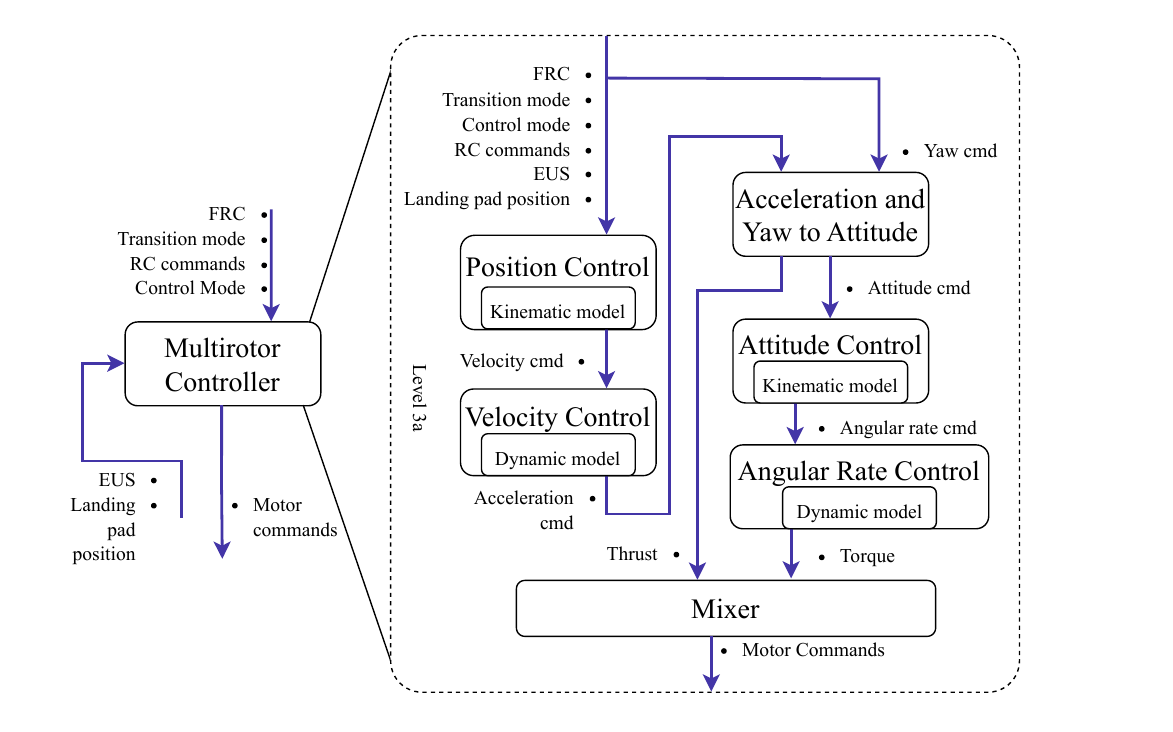}
			\label{fig:Control_struct_L3}	
		}
    \caption{\small (a) Control structure of the autopilot block. (b) Control structure of the multirotor controller block.}	
		\label{fig:Control_struct_L2_L3}
	\end{center}
\end{figure*}

\subsection{Control Structure} 
The control structure for the system is illustrated in~\autoref{fig:Control_struct}. It encompasses all the components described in Section~\ref{sec:system_description}. As indicated, the remote pilot selects the autonomy mode (enabled or disabled) and specifies waypoints to the GCS. The GCS, in turn, provides feedback on the autonomy mode status and estimated UAV states (EUS). The GCS transmits the mission parameters to the guidance controller and receives updates on EUS and sensor data. Additionally, the autonomy mode selected by remote pilot is transmitted to the guidance controller. The guidance controller generates the trajectory to be followed by the VTOL UAV and transmits flight reference commands, flight mode, and control mode to the autopilot. The autopilot then provides feedback on the EUS to the guidance controller, allowing for updates to the reference commands as needed. The autopilot calculates motor actuation commands for multirotor operations and control surface commands for fixed-wing operations. These commands are transmitted to the VTOL UAV’s actuators, which, in turn, provide sensor data feedback to close the control loop.

\begin{figure}[h]
    \centering
    \includegraphics[width=\linewidth]{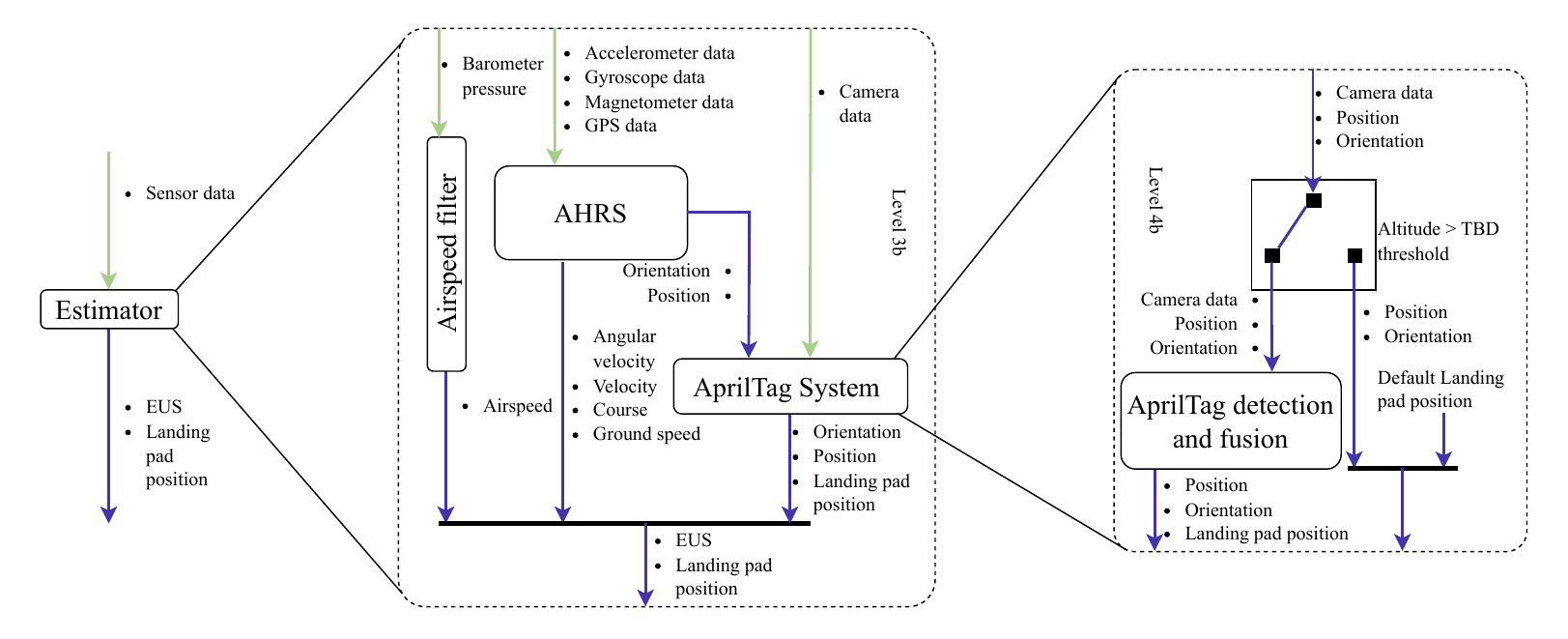}
    \caption{\small Control structure of the Estimator and AprilTag System. The solid horizontal line indicates a mux. }
    \label{fig:Control_struct_L23}
\end{figure}

We expand the control loop to two more level, illustrated in~\autoref{fig:Control_struct_L2_L3} and~\autoref{fig:Control_struct_L23}. The details of the control structure blocks are as follows.
\begin{itemize}
\begin{minipage}[t]{\linewidth}
\setlength{\leftskip}{1.5em}
    \item [Level 2]: This level expands the autopilot block, which includes the multirotor controller, fixed-wing controller, tilt scheduler, and scheduler (Fig.~\ref{fig:Control_struct_L2}). The tilt scheduler is responsible for altering the thrust vector of the motors, which is essential in transition. The scheduler filters motor actuation and control surface commands based on the current flight mode (hover, transition, fixed-wing, or back-transition) and sends these commands to the host VTOL UAV actuators.
    \item [Level 3a]: At this level, the multirotor controller block is expanded to include multiple sub-controllers such as position, velocity, attitude, and angular rate controllers (Fig.~\ref{fig:Control_struct_L3}). The multirotor controller block forms a cascaded control structure in which outer-loop controllers provide reference signals to inner-loop controllers. The multirotor controller generates motor actuation commands to track reference signals from the guidance controller.
    \item [Level 3b]: This level expands the estimator block, which consists of multiple sensor processing modules, including the airspeed filter, altitude and heading reference system (AHRS), GPS processing, and the AprilTag system (Fig.~\ref{fig:Control_struct_L23}). The AprilTag system incorporates the vision processing block, which processes high-resolution visual data for tasks such as landing pad detection and localization. 
    \item [Level 4b]: At this level, the AprilTag block is further expanded to include a switching system and an AprilTag detection and fusion system (Fig.~\ref{fig:Control_struct_L23}). The AprilTag detection and fusion block is kept generalized to facilitate a system-level analysis without delving into specific algorithms used in its implementation.
\end{minipage}
\end{itemize}


\subsection{Identify Unsafe Control Actions}
Next, we perform the STPA step 3, identifying the unsafe control actions (UCAs) for the
multirotor controller and  AprilTag system, indicated in~\autoref{tab:Unsafe_Control_Action}.
The control actions, landing pad position and motor commands can turn unsafe under four broad categories: i) not providing causes hazard, ii) providing causes hazards, iii) too early, too late or out-of-order, and iv) stopped too soon, applied too long. 
\begin{table}[h]
\caption{Identified unsafe control action leading to corresponding hazards for Level 3a and Level 4b control structures, multirotor controller and AprilTag System, respectively.}
\label{tab:Unsafe_Control_Action}
\resizebox{\columnwidth}{!}{%
\begin{tabular}{l|l|l|l|l}
\rowcolor[HTML]{FFFFC7} 
\multicolumn{1}{c|}{\cellcolor[HTML]{FFFFC7}Control Action} &
  \multicolumn{1}{c|}{\cellcolor[HTML]{FFFFC7}\begin{tabular}[c]{@{}c@{}}Not providing\\ causes hazard\end{tabular}} &
  \multicolumn{1}{c|}{\cellcolor[HTML]{FFFFC7}\begin{tabular}[c]{@{}c@{}}Providing \\ causes hazard\end{tabular}} &
  \multicolumn{1}{c|}{\cellcolor[HTML]{FFFFC7}\begin{tabular}[c]{@{}c@{}}Too early, too late,\\ out-of-order\end{tabular}} &
  \multicolumn{1}{c}{\cellcolor[HTML]{FFFFC7}\begin{tabular}[c]{@{}c@{}}Stopped too soon,\\ applied too long\end{tabular}} \\ \hline
Landing pad position &
  \begin{tabular}[c]{@{}l@{}}UCA-1: AprilTag System does\\ not provide vision-based\\ landing pad position when the \\ entire fiducial marker is visible\\ after take-off or during\\ landing {[}H-3,H-4,H-6{]}\end{tabular} &
  \begin{tabular}[c]{@{}l@{}}UCA-2: AprilTag System provides\\ vision-based landing pad position\\ when the fiducial marker is \\ out of sight when taking \\ off or landing {[}H-3, H-4, H-6{]}\end{tabular} &
  \begin{tabular}[c]{@{}l@{}}UCA-3: AprilTag System provides\\ out-of-order vision-based landing\\ pad position when the fiducial\\ marker is visible during\\ take-off or landing {[}H-3, H-4, H-6{]}\end{tabular} &
  \begin{tabular}[c]{@{}l@{}}UCA-4: AprilTag System stops \\ providing the vision-based \\ landing pad position data \\ during the landing {[}H-4{]}\end{tabular}
  \\ \hline
    Motor commands &
  \begin{tabular}[c]{@{}l@{}}UCA-5: Multirotor Controller does not \\ provide motor actuation commands \\ during take-off or landing {[}H-1{]}-{[}H-6{]}\end{tabular} &
  \begin{tabular}[c]{@{}l@{}}UCA-6: Multirotor Controller provides \\ motor actuation commands which \\ deviates from reference trajectory \\ during take-off or landing {[}H-1{]}-{[}H-6{]} \\
  UCA-7: Multirotor Controller continues \\ to provide motor actuation commands \\ after landing {[}H-1, H-4{]}
  \end{tabular} &
  \multicolumn{2}{l}{\begin{tabular}[c]{@{}l@{}}UCA-8: Multirotor Controller provides delayed motor commands \\ during take-off or landing {[}H-1{]}-{[}H-6{]}\end{tabular}}
\end{tabular}%
}
\end{table}

\section{Loss Scenarios and Mitigation Strategies}
Any UCA occurs under four classes of scenarios: Class 1 - the feedback or input is good, but the control action generated is unsafe, Class 2 - both the feedback or input are unsafe and control actions are unsafe, Class 3 - the feedback or input and the generated control action are good, but the control actions become unsafe in the control path, and Class 4 - the generated control action is good but the controlled process produces an unsafe action.
For the unsafe control actions in~\autoref{tab:Unsafe_Control_Action},
we identify two classes of the loss scenarios.
The first class of identified scenarios is described in~\autoref{tab:Class_1_loss_scenarios}.

\begin{table}[h]
\caption{Class 1 loss scenarios for the identified unsafe control actions. }
\label{tab:Class_1_loss_scenarios}
\resizebox{\columnwidth}{!}{%
\begin{tabular}{l|l|l|l|l}
\rowcolor[HTML]{FFFFC7} 
\multicolumn{1}{c|}{\cellcolor[HTML]{FFFFC7}} &
  \multicolumn{1}{c|}{\cellcolor[HTML]{FFFFC7}\begin{tabular}[c]{@{}c@{}}Not providing\\ causes hazard\end{tabular}} &
  \multicolumn{1}{c|}{\cellcolor[HTML]{FFFFC7}\begin{tabular}[c]{@{}c@{}}Providing \\ causes hazard\end{tabular}} &
  \multicolumn{1}{c|}{\cellcolor[HTML]{FFFFC7}\begin{tabular}[c]{@{}c@{}}Too early, too late,\\ out-of-order\end{tabular}} &
  \multicolumn{1}{c}{\cellcolor[HTML]{FFFFC7}\begin{tabular}[c]{@{}c@{}}Stopped too soon,\\ applied too long\end{tabular}} \\ \hline
\cellcolor[HTML]{FFCE93}\begin{tabular}[c]{@{}l@{}}Control Action: \\ Landing pad position, \\ Motor commands\end{tabular} &
  \begin{tabular}[c]{@{}l@{}}UCA-1: AprilTag System does\\ not provide vision-based\\ landing pad position when the \\ entire fiducial marker is visible\\ after take-off or during\\ landing {[}H-3,H-4,H-6{]}\\
  UCA-5: Multirotor Controller does not \\ provide motor actuation commands \\ during take-off or landing {[}H-1{]}-{[}H-6{]}
  \end{tabular} &
  \begin{tabular}[c]{@{}l@{}}UCA-2: AprilTag System provides\\ vision-based landing pad position\\ when the fiducial marker is \\ out of sight when taking \\ off or landing {[}H-3, H-4, H-6{]} \\
  UCA-6: Multirotor Controller provides \\ motor actuation commands which \\ deviates from reference trajectory \\ during take-off or landing {[}H-1{]}-{[}H-6{]}
  \end{tabular} &
  \begin{tabular}[c]{@{}l@{}}UCA-3: AprilTag System provides\\ out-of-order vision-based landing\\ pad position when the fiducial\\ marker is visible during\\ take-off or landing {[}H-3, H-4, H-6{]}\end{tabular} &
  \begin{tabular}[c]{@{}l@{}}
  UCA-4: AprilTag System stops \\ providing the vision-based \\ landing pad position data \\ during the landing {[}H-4{]} \\
  UCA-8: Multirotor Controller provides \\ delayed motor actuation commands \\ during take-off or landing {[}H-1{]}-{[}H-6{]}
  \end{tabular} \\ \hline
\cellcolor[HTML]{FFCE93}Class 1 Scenario &
  \begin{tabular}[c]{@{}l@{}}1. AprilTag system believes the\\ altitude  is below the \\ threshold altitude (UCA-1) \\
  2. Algorithms in Multirotor Controller \\ fail to solve sub-problems (e.g. control \\ allocation) to provide motor actuation \\ commands (UCA-5)\end{tabular} &
  \begin{tabular}[c]{@{}l@{}} 1. AprilTag system believes the\\ UAV altitude is above the \\ threshold altitude (UCA-2). \\
  2. The PID model (gains) of the \\
  multirotor controller is updated \\
  incorrectly due to change in \\
  operating (trim) condition (UCA-6). \end{tabular} &
  \begin{tabular}[c]{@{}l@{}}AprilTag detection and fusion \\ generated out-of-order landing \\ pad position data. \end{tabular} &
  \begin{tabular}[c]{@{}l@{}}
  1. AprilTag system believes the\\ altitude  is below the \\ threshold altitude (UCA-4) \\
  2. Algorithms in Multirotor Controller \\ provide motor actuation commands \\ too late, and the previous \\ commands are applied too long (UCA-8)
  \end{tabular}
\end{tabular}%
}
\end{table}
\begin{table}[h]
\caption{Class 2 loss scenarios for the identified unsafe control actions. }
\label{tab:Class_2_loss_scenarios}
\resizebox{\columnwidth}{!}{%
\begin{tabular}{l|l|l|l|l}
\rowcolor[HTML]{FFFFC7} 
\multicolumn{1}{c|}{\cellcolor[HTML]{FFFFC7}} &
  \multicolumn{1}{c|}{\cellcolor[HTML]{FFFFC7}\begin{tabular}[c]{@{}c@{}}Not providing\\ causes hazard\end{tabular}} &
  \multicolumn{1}{c|}{\cellcolor[HTML]{FFFFC7}\begin{tabular}[c]{@{}c@{}}Providing \\ causes hazard\end{tabular}} &
  \multicolumn{1}{c|}{\cellcolor[HTML]{FFFFC7}\begin{tabular}[c]{@{}c@{}}Too early, too late,\\ out-of-order\end{tabular}} &
  \multicolumn{1}{c}{\cellcolor[HTML]{FFFFC7}\begin{tabular}[c]{@{}c@{}}Stopped too soon,\\ applied too long\end{tabular}} \\ \hline
\cellcolor[HTML]{FFCE93}\begin{tabular}[c]{@{}l@{}}Control Action: \\ Landing pad position, \\ Motor commands \end{tabular} &
  \begin{tabular}[c]{@{}l@{}}UCA-1: AprilTag System does\\ not provide vision-based\\ landing pad position when the \\ entire fiducial marker is visible\\ after take-off or during\\ landing {[}H-3,H-4,H-6{]}\end{tabular} &
  \begin{tabular}[c]{@{}l@{}}UCA-2: AprilTag System provides\\ vision-based landing pad position\\ when the fiducial marker is \\ out of sight when taking \\ off or landing {[}H-3, H-4, H-6{]} \\
  UCA-7: Multirotor Controller  \\ continues to provide motor \\ actuation commands  after \\ landing {[}H-1{]},{[}H-4{]}
  \end{tabular} &
  \begin{tabular}[c]{@{}l@{}}UCA-3: AprilTag System provides\\ out-of-order vision-based landing\\ pad position when the fiducial\\ marker is visible during\\ take-off or landing {[}H-3, H-4, H-6{]}\end{tabular} &
  \begin{tabular}[c]{@{}l@{}}UCA-4: AprilTag System stops \\ providing the vision-based \\ landing pad position data \\ during the landing {[}H-4{]}\end{tabular} \\ \hline
\cellcolor[HTML]{FFCE93}Class 2 Scenario &
  \begin{tabular}[c]{@{}l@{}} 1. The camera data is corrupted\\ due to occlusion of the\\ fiducial marker. \\ 2. Insufficient lighting \\ conditions for AprilTag \\ detection system. \end{tabular} &
  \begin{tabular}[c]{@{}l@{}}
  1. Additional neighboring \\ fiducial marker of \\ different sizes causes a \\ different landing pad\\  position data (UCA-2). \\ 
  2. The EUS indicates the VTOL \\ UAV is below the reference \\ altitude (UCA-7).
  \end{tabular} &
   &
  \begin{tabular}[c]{@{}l@{}}1. The camera data is corrupted\\ due to occlusion of the\\ fiducial marker. \\ 2. Insufficient lighting \\ conditions for AprilTag \\ detection system.\end{tabular}
\end{tabular}%
}
\end{table}

\textbf{Solution}: The Class 1 scenario corresponding to the action of landing pad location is common to both UCA-1 and UCA-4. A major cause of the Class 1 for the landing pad location loss scenario arises from the altitude state input, which leads to the control action behaving unsafely. For the motor commands, the loss scenario primarily occurs when the tuned proportional integral and derivation (PID) gains are operating in different conditions.
These conditions can either be model mismatch or external disturbances (wind), referred to as trim conditions.  To mitigate this issue, we propose the following control strategies:
\begin{itemize}
    \item Introduce a secondary altitude measurement as an additional input to the AprilTag system. A potential solution could involve using an infrared (IR) sensor in conjunction with an IR beacon on the landing pad. This setup can aid the switching mechanism within the AprilTag system, enhancing its reliability (UCA-1, UCA-2, UCA-4).
    \item Optimize the camera's image capture rate and adjust the processing resolution to prevent bottlenecks and mitigate the risk of out-of-order image generation by the AprilTag system (UCA-3).
    \item Verify and optimize software for providing motor actuation commands (UCA-5, UCA-8).
    \item Append an adaptive controller that accounts for model uncertainties and external disturbances (UCA-6).
\end{itemize}

\smallskip 
\noindent
The last solution improves the robustness of the multirotor controller by adaptation.
Unlike kinematic models (position and attitude), dynamic models (velocity and angular rate) can become uncertain due to external disturbances such as wind and drag,
implying that the process models used for controller design may differ from the actual models.
To address this, adaptive control methods such as $\mathcal{L}_{1}$ adaptive control~\cite{hovakimyan2010ℒ1,wu2023mathcal} can be utilized to improve the control performance.
The second class of identified scenarios is described in~\autoref{tab:Class_2_loss_scenarios}.

\textbf{Solution}: Similar to the Class 1 scenarios, the Class 2 scenarios corresponding to landing pad position share common causes for UCA-1 and UCA-4.
A primary contributor to the Class 2 loss scenario corresponding to landing pad location is the reliance on a single fiducial marker and the absence of a comprehensive tagging system for individual VTOLs.
Based on the identified scenarios, we propose the following control and feedback mitigation strategies:
\begin{itemize}
    \item Ensure the remote pilot performs an occlusion check on the fiducial marker before initiating the mission to confirm visibility (UCA-1).
    \item Conduct the mission only under predefined (TBD) lighting conditions to avoid detection failures caused by low visibility (UCA-1). 
    \item Deploy multiple fiducial markers to provide redundancy, enhancing reliability and ensuring robust landing pad detection (UCA-2, UCA-4).  
    \item Integrate a tagging system within the AprilTag detection and fusion system to uniquely identify landing pad locations, allowing precise recognition even in complex environments (UCA-2, UCA-4).
    \item Utilize a secondary altitude measurement to refine the EUS (UCA-7).
\end{itemize}

\begin{wrapfigure}{l}{0.35\textwidth} 
    \centering
    \includegraphics[width=0.33\textwidth]{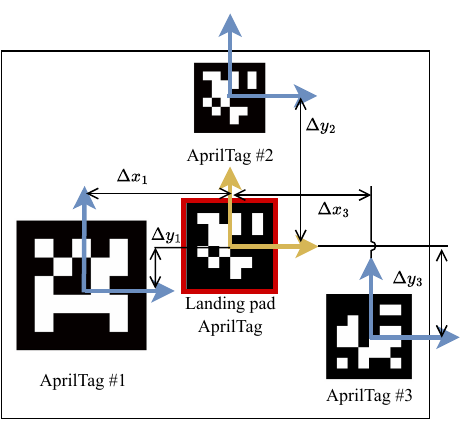}
    \caption{\small Multiple fiducial markers (AprilTags) with varying size and location for enable robust landing and take-off of VTOL UAV. }
    \label{fig:multiple_AT}
\end{wrapfigure}

\noindent
The first solution corresponds to a checklist that must be completed by the remote pilot before initiating the mission. The second solution is dynamic and requires experimental validation to determine the optimal lighting conditions for reliable operation. The third and fourth solutions necessitate modifications to the underlying AprilTag detection and fusion system to accommodate multiple fiducial markers. A visual representation of the third and fourth solutions is provided in~\autoref{fig:multiple_AT}. While existing multi-fiducial marker setups have been explored, the proposed approach utilizes markers of varying sizes, which enhances robustness by ensuring reliable detection even if some markers are occluded. 

Additionally, since the locations of the auxiliary fiducial markers are predefined and stored within the VTOL UAV autopilot, the system can achieve greater accuracy in identifying the landing pad location. This is accomplished by averaging or applying weighted averaging to the measurements from multiple markers, thereby improving precision in the detection of the landing site. By integrating these strategies, the VTOL UAV can achieve improved safety and precision during critical flight phases, particularly take-off and landing. This foundation paves the way for future advancements in vision-based navigation systems, ensuring their adaptability to complex and uncertain operational scenarios.

\section{Conclusion}
This work addresses the critical safety challenges of VTOL UAV operations during take-off and landing by integrating vision-based systems with System-Theoretic Process Analysis (STPA). Through detailed hazard identification and control strucutre analysis, we systematically addressed unsafe control actions associated with the vision-based landing system and multirotor controller. The proposed multi-fiducial marker approach enhances the robustness of landing pad detection under occlusion and dynamic lighting conditions, while additional redundancy measures and altitude state integration improve system reliability. Similarly, we propose to add adaptive mechanism and modification of multirotor controller corresponding to the motor command control action. Experimental insights and design recommendations were provided to refine vision-based sensor fusion mechanisms along and motor control commands to assure safer UAV operations. The findings contribute to the advancement of autonomous VTOL UAV technologies, establishing a framework for future research on safety-critical systems in dynamic environments. Future work involves experimental testing and implementation of the proposed solutions for both the classes of scenarios and extending the STPA analysis for the complete autonomous mission. 

\section{Acknowledgment}
This material is based upon work supported by the National Aeronautics and Space Administration (NASA) under the cooperative agreement 80NSSC20M0229, University Leadership Initiative grant no. 80NSSC22M0070 and Air Force Office of Scientific Research (AFOSR) grant no. FA9550-21-1-0411. This work was also supported by the AFOSR ``Certifiable and Self-Supervised Category-Level Tracking” program, Carlone’s NSF CAREER award. Any opinions, findings, conclusions or recommendations expressed in this material are those of the authors and do not necessarily reflect the
views of the sponsors.

\vspace{10pt}

\bibliography{sample}

\end{document}